\documentstyle[11pt,colacl]{article}

\title{Cross-Language Information Retrieval for Technical Documents}
\author{Atsushi Fujii \and Tetsuya Ishikawa \\
University of Library and Information Science \\
1-2 Kasuga Tsukuba 305-8550, JAPAN \\ \smallskip
{\tt \{fujii,ishikawa\}@ulis.ac.jp}}

\begin{document}

\maketitle

\input{psfig.tex}

\newcommand{\etal}{et~al.}
\newcommand{\etaleos}{et~al}
\newcommand{\eq}[1]{(\ref{#1})}

\begin{abstract}
  This paper proposes a Japanese/English cross-language information
  retrieval (CLIR) system targeting technical documents.  Our system
  first translates a given query containing technical terms into the
  target language, and then retrieves documents relevant to the
  translated query. The translation of technical terms is still
  problematic in that technical terms are often compound words, and
  thus new terms can be progressively created simply by combining
  existing base words.  In addition, Japanese often represents
  loanwords based on its phonogram.  Consequently, existing
  dictionaries find it difficult to achieve sufficient coverage. To
  counter the first problem, we use a compound word translation
  method, which uses a bilingual dictionary for base words and
  collocational statistics to resolve translation ambiguity. For the
  second problem, we propose a transliteration method, which
  identifies phonetic equivalents in the target language. We also show
  the effectiveness of our system using a test collection for CLIR.
\end{abstract}

\section{Introduction}
\label{sec:introduction}

Cross-language information retrieval (CLIR), where the user presents
queries in one language to retrieve documents in another language, has
recently been one of the major topics within the information retrieval
community.  One strong motivation for CLIR is the growing number of
documents in various languages accessible via the Internet.  Since
queries and documents are in different languages, CLIR requires a
translation phase along with the usual monolingual retrieval phase.
For this purpose, existing CLIR systems adopt various techniques
explored in natural language processing (NLP) research. In brief,
bilingual dictionaries, corpora, thesauri and machine translation (MT)
systems are used to translate queries or/and documents.

In this paper, we propose a Japanese/English CLIR system for technical
documents, focusing on translation of technical terms. Our purpose
also includes integration of different components within one
framework.  Our research is partly motivated by the ``NACSIS'' test
collection for IR systems~\cite{kando:bcs-irsg-98}\footnote{\scriptsize\tt
{http://www.rd.nacsis.ac.jp/\~{}ntcadm/index-en.html}}, which consists
of Japanese queries and Japanese/English abstracts extracted from
technical papers (we will elaborate on the NACSIS collection in
Section~\ref{sec:evaluation}).  Using this collection, we investigate
the effectiveness of each component as well as the overall performance
of the system.

As with MT systems, existing CLIR systems still find it difficult to
translate technical terms and proper nouns, which are often unlisted
in general dictionaries.  Since most CLIR systems target newspaper
articles, which are comprised mainly of general words, the problem
related to unlisted words has been less explored than other CLIR
subtopics (such as resolution of translation ambiguity). However,
Pirkola~\shortcite{pirkola:sigir-98}, for example, used a subset of
the TREC collection related to health topics, and showed that
combination of general and domain specific (i.e., medical)
dictionaries improves the CLIR performance obtained with only a
general dictionary.  This result shows the potential contribution of
technical term translation to CLIR.  At the same time, note that even
domain specific dictionaries do not exhaustively list possible
technical terms.  We classify problems associated with technical term
translation as given below:
\begin{enumerate}
  \def\labelenumi{(\theenumi)}
\item \label{enu:cw} technical terms are often compound word, which
  can be progressively created simply by combining multiple existing
  morphemes (``base words''), and therefore it is not entirely
  satisfactory to exhaustively enumerate newly emerging terms in
  dictionaries,
\item \label{enu:kana} Asian languages often represent loanwords based
  on their special phonograms (primarily for technical terms and
  proper nouns), which creates new base words progressively (in the
  case of Japanese, the phonogram is called {\it katakana}).
\end{enumerate}
To counter problem~(\ref{enu:cw}), we use the compound word
translation method we proposed~\cite{fujii:iccpol-99}, which selects
appropriate translations based on the probability of occurrence of
each combination of base words in the target language.  For
problem~(\ref{enu:kana}), we use
``transliteration''~\cite{chen:coling-acl-98,knight:cl-98,wan:coling-acl-98}.
Chen~\etal~\shortcite{chen:coling-acl-98} and Wan and
Verspoor~\shortcite{wan:coling-acl-98} proposed English-Chinese
transliteration methods relying on the property of the Chinese
phonetic system, which cannot be directly applied to transliteration
between English and Japanese. Knight and
Graehl~\shortcite{knight:cl-98} proposed a Japanese-English
transliteration method based on the mapping probability between
English and Japanese {\it katakana\/} sounds. However, since their
method needs large-scale phoneme inventories, we propose a simpler
approach using surface mapping between English and {\it katakana\/}
characters, rather than sounds.

Section~\ref{sec:overview} overviews our CLIR system, and
Section~\ref{sec:translation} elaborates on the translation module
focusing on compound word translation and
transliteration. Section~\ref{sec:evaluation} then evaluates the
effectiveness of our CLIR system by way of the standardized IR
evaluation method used in TREC programs.

\section{System Overview}
\label{sec:overview}

Before explaining our CLIR system, we classify existing CLIR into
three approaches in terms of the implementation of the translation
phase. The first approach translates queries into the document
language~\cite{ballesteros:sigir-98,carbonell:ijcai-97,davis:sigir-97,fujii:iccpol-99,hull:sigir-96,kando:iral-98,okumura:lrec-tlim-ws-98},
while the second approach translates documents into the query
language~\cite{gachot:sigir-ws-96,oard:trec-97}.  The third approach
transfers both queries and documents into an interlingual
representation: bilingual thesaurus
classes~\cite{mongar:tis-69,salton:jasis-70,sheridan:sigir-96} and
language-independent vector space
models~\cite{carbonell:ijcai-97,dumais:sigir-ws-96}.  We prefer the
first approach, the ``query translation'', to other approaches because
(a) translating all the documents in a given collection is expensive,
(b) the use of thesauri requires manual construction or bilingual {\em
comparable\/} corpora, (c) interlingual vector space models also need
comparable corpora, and (d) query translation can easily be combined
with existing IR engines and thus the implementation cost is low.  At
the same time, we concede that other CLIR approaches are worth further
exploration.

\begin{figure}[htbp]
  \begin{center}
    \leavevmode
    \psfig{file=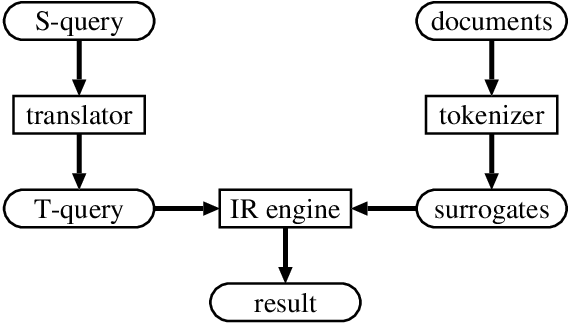,height=1.5in}
  \end{center}
  \caption{The overall design of our CLIR system}
  \label{fig:system}
\end{figure}

Figure~\ref{fig:system} depicts the overall design of our CLIR system,
where most components are the same as those for monolingual
IR, excluding ``translator''.

First, ``tokenizer'' processes \mbox{``documents''} in a given
collection to produce an inverted file (``surrogates''). Since our
system is bidirectional, tokenization differs depending on the target
language.  In the case where documents are in English, tokenization
involves eliminating stopwords and identifying root forms for
inflected words, for which we used
``WordNet''~\cite{miller:techrep-93}.  On the other hand, we segment
Japanese documents into lexical units using the ``ChaSen''
morphological analyzer~\cite{matsumoto:chasen-97} and discard
stopwords.  In the current implementation, we use word-based uni-gram
indexing for both English and Japanese documents. In other words,
compound words are decomposed into base words in the surrogates. Note
that indexing and retrieval methods are theoretically independent of
the translation method.

Thereafter, the ``translator'' processes a query in the source
language (``S-query'') to output the translation (``T-query'').
T-query can consist of more than one translation, because multiple
translations are often appropriate for a single technical term.

Finally, the ``IR engine'' computes the similarity between T-query and
each document in the surrogates based on the vector space
model~\cite{salton:83}, and sorts document according to the
similarity, in descending order. We compute term weight based on the
notion of TF$\cdot$IDF. Note that T-query is decomposed into base
words, as performed in the document preprocessing.

In Section~\ref{sec:translation}, we will explain the
\mbox{``translator''} in Figure~\ref{fig:system}, which involves
compound word translation and transliteration modules.

\section{Translation Module}
\label{sec:translation}

\subsection{Overview}
\label{subsec:trans_overview}

Given a query in the source language, tokenization is first performed
as for target documents (see Figure~\ref{fig:system}). To put it more
precisely, we use WordNet and ChaSen for English and Japanese queries,
respectively. We then discard stopwords and extract only content
words. Here, ``content words'' refer to both single and compound
words.  Let us take the following query as an example:
\begin{list}{}{}
\item improvement of data mining methods.
\end{list}
For this query, we discard ``of'', to extract ``improvement'' and
``data mining methods''.

Thereafter, we translate each extracted content word
individually. Note that we currently do not consider relation (e.g.
syntactic relation and collocational information) between content
words. If a single word, such as ``improvement'' in the example above,
is listed in our bilingual dictionary (we will explain the way to
produce the dictionary in Section~\ref{subsec:cwt}), we use all
possible translation candidates as query terms for the subsequent
retrieval phase.

Otherwise, compound word translation is performed. In the case of
Japanese-English translation, we consider all possible segmentations
of the input word, by consulting the dictionary. Then, we select such
segmentations that consist of the minimal number of base words. During
the segmentation process, the dictionary derives all possible
translations for base words. At the same time, transliteration is
performed whenever {\it katakana\/} sequences unlisted in the
dictionary are found.  On the other hand, in the case of
English-Japanese translation, transliteration is applied to any
unlisted base word (including the case where the input English word
consists of a single base word).  Finally, we compute the probability
of occurrence of each combination of base words in the target
language, and select those with greater probabilities, for both
Japanese-English and English-Japanese translations.

\subsection{Compound Word Translation}
\label{subsec:cwt}

This section briefly explains the compound word translation method we
previously proposed~\cite{fujii:iccpol-99}. This method translates
input compound words on a word-by-word basis, maintaining the word
order in the source language\footnote{A preliminary study showed that
approximately 95\% of compound technical terms defined in a bilingual
dictionary maintain the same word order in both source and target
languages.}.  The formula for the source compound word and one
translation candidate are represented as below.
\begin{eqnarray*}
    S & = & s_{1}, s_{2}, \ldots, s_{n} \\
    T & = & t_{1}, t_{2}, \ldots, t_{n}
\end{eqnarray*}
Here, $s_{i}$ and $t_{i}$ denote $i$-th base words in source and
target languages, respectively. Our task, i.e., to select $T$ which
maximizes $P(T|S)$, is transformed into Equation~\eq{eq:trans_model}
through use of the Bayesian theorem.
\begin{equation}
  \label{eq:trans_model}
  \arg\max_{T}P(T|S) = \arg\max_{T}P(S|T)\cdot P(T)
\end{equation}
$P(S|T)$ and $P(T)$ are approximated as in Equation~\eq{eq:approx},
which has commonly been used in the recent statistical NLP
research~\cite{church:cl-93}.
\begin{equation}
  \label{eq:approx}
  \begin{array}{lll}
    P(S|T) & \approx & {\displaystyle \prod_{i=1}^{n}P(s_{i}|t_{i})} \\
    \noalign{\vskip 1.2ex}
    P(T) & \approx & {\displaystyle \prod_{i=1}^{n-1}P(t_{i+1}|t_{i})}
  \end{array}
\end{equation}

We produced our own dictionary, because conventional dictionaries are
comprised primarily of general words and verbose definitions aimed at
human readers. We extracted 59,533 English/Japanese translations
consisting of {\em two\/} base words from the EDR technical
terminology dictionary, which contains about 120,000 translations
related to the information processing field~\cite{edr-techdic:95}, and
segment Japanese entries into two parts\footnote{The number of base
words can easily be identified based on English words, while Japanese
compound words lack lexical segmentation.}. For this purpose, simple
heuristic rules based mainly on Japanese character types (i.e., {\it
kanji}, {\it katakana}, {\it hiragana}, alphabets and other characters
like numerals) were used. Given the set of compound words where
Japanese entries are segmented, we correspond English-Japanese base
words on a word-by-word basis, maintaining the word order between
English and Japanese, to produce a Japanese-English/English-Japanese
base word dictionary.  As a result, we extracted 24,439 Japanese base
words and 7,910 English base words from the EDR dictionary.  During
the dictionary production, we also count the collocational frequency
for each combination of $s_{i}$ and $t_{i}$, in order to estimate
$P(s_{i}|t_{i})$. Note that in the case where $s_{i}$ is {\em
transliterated\/} into $t_{i}$, we use an arbitrarily predefined value
for $P(s_{i}|t_{i})$.  For the estimation of $P(t_{i+1}|t_{i})$, we
use the word-based bi-gram statistics obtained from target language
corpora, i.e., ``documents'' in the collection (see
Figure~\ref{fig:system}).

\subsection{Transliteration}
\label{subsec:translit}

Figure~\ref{fig:katakana} shows example correspondences between
English and (romanized) {\it katakana\/} words, where we insert
hyphens between each {\it katakana\/} character for enhanced
readability.  The basis of our transliteration method is analogous to
that for compound word translation described in
Section~\ref{subsec:cwt}.  The formula for the source word and one
transliteration candidate are represented as below.
\begin{eqnarray*}
    S & = & s_{1}, s_{2}, \ldots, s_{n} \\
    T & = & t_{1}, t_{2}, \ldots, t_{n}
\end{eqnarray*}
However, unlike the case of compound word translation, $s_{i}$ and
$t_{i}$ denote $i$-th ``symbols'' (which consist of one or more
letters), respectively. Note that we consider only such $T$'s that are
indexed in the inverted file, because our transliteration method often
outputs a number of incorrect words with great probabilities.  Then,
we compute $P(T|S)$ for each $T$ using Equations~\eq{eq:trans_model}
and \eq{eq:approx} (see Section~\ref{subsec:cwt}), and select $k$-best
candidates with greater probabilities.  The crucial content here is
the way to produce a bilingual dictionary for {\em symbols}.  For this
purpose, we used approximately 3,000 {\it katakana\/} entries and
their English translations listed in our base word dictionary.  To
illustrate our dictionary production method, we consider
Figure~\ref{fig:katakana} again. Looking at this figure, one may
notice that the first letter in each {\it katakana\/} character tends
to be contained in its corresponding English word. However, there are
a few exceptions. A typical case is that since Japanese has no
distinction between ``L'' and ``R'' sounds, the two English sounds
collapse into the same Japanese sound. In addition, a single English
letter corresponds to multiple {\it katakana\/} characters, such as
``x'' to ``{\it ki-su\/}'' in \mbox{``$<$text, {\it
te-ki-su-to\/}$>$''}. To sum up, English and romanized {\it
katakana\/} words are not exactly identical, but {\em similar\/} to
each other.

\begin{figure}[htbp]
  \def\baselinestretch{1}
  \begin{center}
    \leavevmode
    \small
    \begin{tabular}{|l|l|} \hline
      {\hfill\centering English \hfill} & {\hfill\centering {\it
      katakana\/}\hfill} \\ \hline
      system & {\it shi-su-te-mu\/} \\
      mining & {\it ma-i-ni-n-gu\/} \\
      data & {\it dee-ta\/} \\
      network & {\it ne-tto-waa-ku\/} \\
      text & {\it te-ki-su-to\/} \\
      collocation & {\it ko-ro-ke-i-sho-n\/} \\ \hline
    \end{tabular}
    \caption{Examples of English-{\it katakana\/} correspondence}
    \label{fig:katakana}
  \end{center}
\end{figure}

\begin{figure}[htbp]
  \begin{center}
    \leavevmode
    \psfig{file=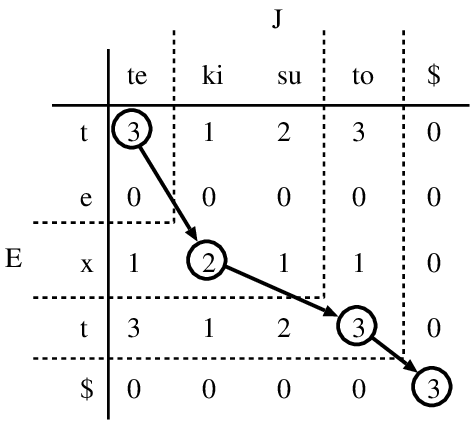,height=2in}
  \end{center}
  \caption{An example matrix for English-Japanese symbol matching
  (arrows denote the best path)}
  \label{fig:matrix}
\end{figure}

We first manually define the similarity between the English letter $e$
and the first romanized letter for each {\it katakana\/} character
$j$, as shown in Table~\ref{tab:katakana}. In this table,
``phonetically similar'' letters refer to a certain pair of letters,
such as ``L'' and ``R''\footnote{We identified approximately twenty
pairs of phonetically similar letters.}.  We then consider the
similarity for any possible combination of letters in English and
romanized {\it katakana\/} words, which can be represented as a
matrix, as shown in Figure~\ref{fig:matrix}. This figure shows the
similarity between letters in \mbox{``$<$text, {\it
te-ki-su-to\/}$>$''}.  We put a dummy letter ``\$'', which has a
positive similarity only to itself, at the end of both English and
{\it katakana\/} words.  One may notice that matching plausible
symbols can be seen as finding the path which maximizes the total
similarity from the first to last letters.  The best path can easily
be found by, for example, Dijkstra's algorithm~\cite{dijkstra:nm-59}.
From Figure~\ref{fig:matrix}, we can derive the following
correspondences: \mbox{``$<$te, {\it te\/}$>$''}, \mbox{``$<$x, {\it
ki-su\/}$>$''} and \mbox{``$<$t, {\it to\/}$>$''}.  The resultant
correspondences contain 944 Japanese and 790 English symbol types,
from which we also estimated $P(s_{i}|t_{i})$ and $P(t_{i+1}|t_{i})$.

As can be predicted, a preliminary experiment showed that our
transliteration method is not accurate when compared with a word-based
translation.  For example, the Japanese word \mbox{``{\it
re-ji-su-ta}~(register)''} is transliterated to ``resister'',
``resistor'' and ``register'', with the probability score in
descending order.  However, combined with the compound word
translation, irrelevant transliteration outputs are expected to be
discarded.  For example, a compound word like ``{\it re-ji-su-ta\/}
{\it tensou\/} {\it gengo\/}~(register transfer language)'' is
successfully translated, given a set of base words ``{\it
tensou\/}~(transfer)'' and ``{\it gengo\/}~(language)'' as a context.

\begin{table}[htbp]
  \def\baselinestretch{1}
  \begin{center}
    \caption{The similarity between English and Japanese letters}
    \medskip
    \leavevmode
    \footnotesize
    \begin{tabular}{|l|c|} \hline
      {\hfill\centering condition \hfill} & {\hfill\centering
      similarity \hfill} \\ \hline\hline
      $e$ and $j$ are identical & 3 \\ \hline
      $e$ and $j$ are phonetically similar & 2 \\ \hline
      both $e$ and $j$ are vowels or consonants & 1 \\ \hline
      otherwise & 0 \\ \hline
    \end{tabular}
    \label{tab:katakana}
  \end{center}
\end{table}

\section{Evaluation}
\label{sec:evaluation}

This section investigates the performance of our CLIR system based on
the TREC-type evaluation methodology: the system outputs 1,000 top
documents, and TREC evaluation software is used to calculate the
recall-precision trade-off and 11-point average precision.

For the purpose of our evaluation, we used the NACSIS test
collection~\cite{kando:bcs-irsg-98}.  This collection consists of 21
Japanese queries and approximately 330,000 documents (in either a
combination of English and Japanese or either of the languages
individually), collected from technical papers published by 65
Japanese associations for various fields.  Each document consists of
the document ID, title, name(s) of author(s), name/date of conference,
hosting organization, abstract and keywords, from which titles,
abstracts and keywords were used for our evaluation. We used as target
documents approximately 187,000 entries where abstracts are in both
English and Japanese.  Each query consists of the title of the topic,
description, narrative and list of synonyms, from which we used only
the description.  Roughly speaking, most topics are related to
electronic, information and control
engineering. Figure~\ref{fig:query} shows example descriptions
(translated into English by one of the authors). Relevance assessment
was performed based on one of the three ranks of relevance, i.e.,
``relevant'', ``partially relevant'' and ``irrelevant''.  In our
evaluation, relevant documents refer to both ``relevant'' and
``partially relevant'' documents\footnote{The result did not
significantly change depending on whether we regarded ``partially
relevant'' as relevant or not.}.

\begin{figure}[htbp]
  \def\baselinestretch{1}
  \begin{center}
    \leavevmode
    \small
    \begin{tabular}{|c|l|} \hline
      ID & {\hfill\centering description\hfill} \\ \hline
      0005 & dimension reduction for clustering \\
      0006 & intelligent information retrieval \\
      0019 & syntactic analysis methods for Japanese \\
      0024 & machine translation systems \\ \hline
    \end{tabular}
    \caption{Example descriptions in the NACSIS query}
    \label{fig:query}
  \end{center}
\end{figure}

\subsection{Evaluation of compound word translation}
\label{subsec:eval_cwt}

\begin{figure}[htbp]
  \begin{center}
    \leavevmode
    \psfig{file=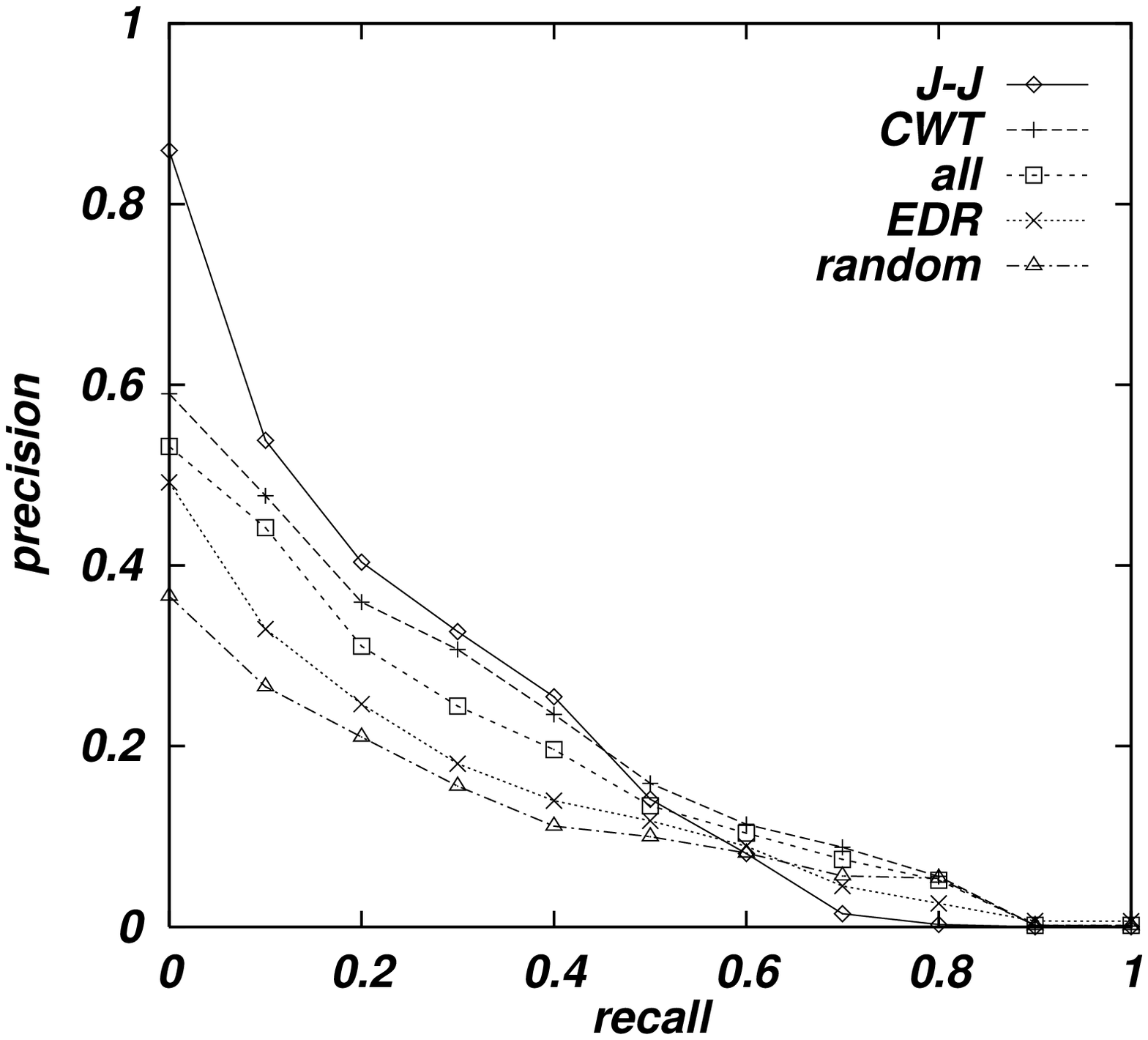,height=2.5in}
  \end{center}
  \caption{Recall-Precision curves for evaluation of compound word translation}
  \label{fig:rp_cwt}
\end{figure}

We compared the following query translation methods:
\begin{enumerate}
  \def\labelenumi{(\theenumi)}
\item a control, in which all possible translations derived from the
  (original) EDR technical terminology dictionary are used as query
  terms (``EDR''),
\item all possible base word translations derived from {\em our\/}
  dictionary are used (``all''),
\item randomly selected $k$ translations derived from our bilingual
  dictionary are used (``random''),
\item $k$-best translations through compound word translation are used
  (``CWT'').
\end{enumerate}
For system~``EDR'', compound words unlisted in the EDR dictionary were
manually segmented so that substrings (shorter compound words or base
words) can be translated.  For both systems ``random'' and ``CWT'', we
arbitrarily set \mbox{$k = 3$}.  Figure~\ref{fig:rp_cwt} and
Table~\ref{tab:ap_cwt} show the recall-precision curve and 11-point
average precision for each method, respectively. In these, ``J-J''
refers to the result obtained by the Japanese-Japanese IR system,
which uses as documents Japanese titles/abstracts/keywords comparable
to English fields in the NACSIS collection. This can be seen as the
upper bound for CLIR performance\footnote{Regrettably, since the
NACSIS collection does not contain English queries, we cannot estimate
the upper bound performance by English-English IR.}.  Looking at these
results, we can conclude that the dictionary production and
probabilistic translation methods we proposed are effective for CLIR.

\begin{table}[htbp]
  \def\baselinestretch{1}
  \begin{center}
    \caption{Comparison of average precision for evaluation of
    compound word translation}
    \medskip
    \leavevmode
    \small
    \begin{tabular}{|c||c|c|} \hline
      & avg. precision & ratio to J-J \\ \hline\hline
      J-J & 0.204 & --- \\ \hline
      CWT & 0.193 & 0.946 \\ \hline
      all & 0.171 & 0.838 \\ \hline
      EDR & 0.130 & 0.637 \\ \hline
      random & 0.116 & 0.569 \\ \hline
    \end{tabular}
    \label{tab:ap_cwt}
  \end{center}
\end{table}

\subsection{Evaluation of transliteration}
\label{subsec:eval_translit}

\begin{figure}[htbp]
  \begin{center}
    \leavevmode
    \psfig{file=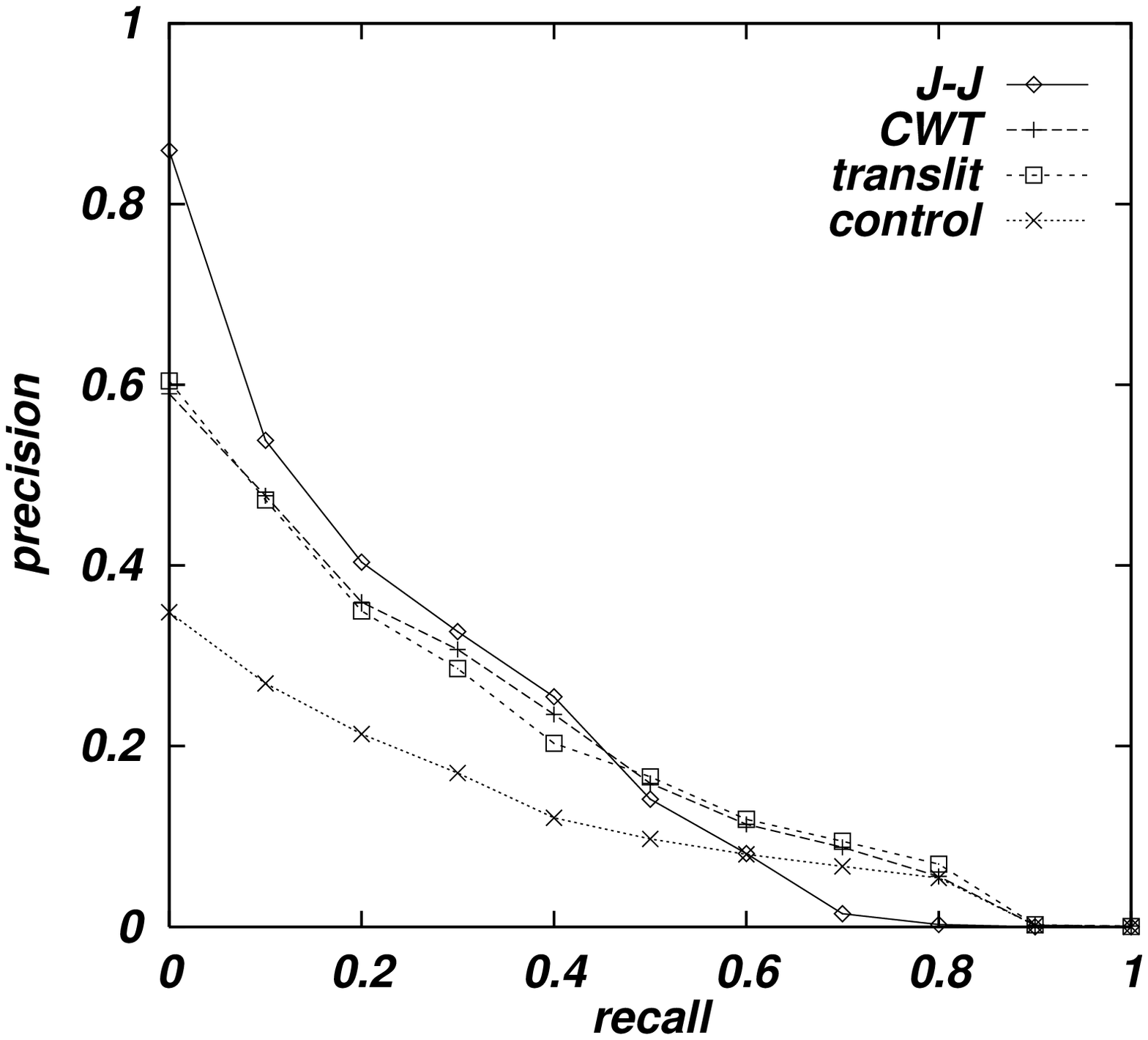,height=2.5in}
  \end{center}
  \caption{Recall-Precision curves for evaluation of transliteration}
  \label{fig:rp_translit}
\end{figure}

In the NACSIS collection, three queries contain {\it katakana\/}
(base) words unlisted in our bilingual dictionary.  Those words are
``{\it ma-i-ni-n-gu\/}~(mining)'' and ``{\it
ko-ro-ke-i-sho-n\/}~(collocation)''. However, to emphasize the
effectiveness of transliteration, we compared the following
extreme cases:
\begin{enumerate}
  \def\labelenumi{(\theenumi)}
\item a control, in which every {\it katakana\/} word is
  discarded from queries (``control''),
\item a case where transliteration is applied to every {\it
  katakana\/} word and top 10 candidates are used
  \mbox{(``translit'')}.
\end{enumerate}
Both cases use system~``CWT'' in Section~\ref{subsec:eval_cwt}.
In the case of ``translit'', we do not use {\it katakana\/} entries
listed in the base word dictionary.  Figure~\ref{fig:rp_translit} and
Table~\ref{tab:ap_translit} show the recall-precision curve and
11-point average precision for each case, respectively. In these,
results for ``CWT'' correspond to those in Figure~\ref{fig:rp_cwt} and
Table~\ref{tab:ap_cwt}, respectively. We can conclude that our
transliteration method significantly improves the baseline performance
(i.e., ``control''), and comparable to word-based translation in terms
of CLIR performance.

An interesting observation is that the use of transliteration is
robust against {\em typos} in documents, because a number of similar
strings are used as query terms. For example, our transliteration
method produced the following strings for ``{\it
ri-da-ku-sho-n\/}~(reduction)'':
\begin{quote}
  riduction,~~redction,~~redaction,~~reduction.
\end{quote}
All of these words are effective for retrieval, because they are
contained in the target documents.

\begin{table}[htbp]
  \def\baselinestretch{1}
  \begin{center}
    \caption{Comparison of average precision for evaluation of
    transliteration}
    \medskip
    \leavevmode
    \small
    \begin{tabular}{|c||c|c|} \hline
      & avg. precision & ratio to J-J \\ \hline\hline
      J-J & 0.204 & --- \\ \hline
      CWT & 0.193 & 0.946 \\ \hline
      translit & 0.193 & 0.946 \\ \hline
      control & 0.115 & 0.564 \\ \hline
    \end{tabular}
    \label{tab:ap_translit}
  \end{center}
\end{table}

\subsection{Evaluation of the overall performance}
\label{subsec:eval_overall}

\begin{figure}[htbp]
  \begin{center}
    \leavevmode
    \psfig{file=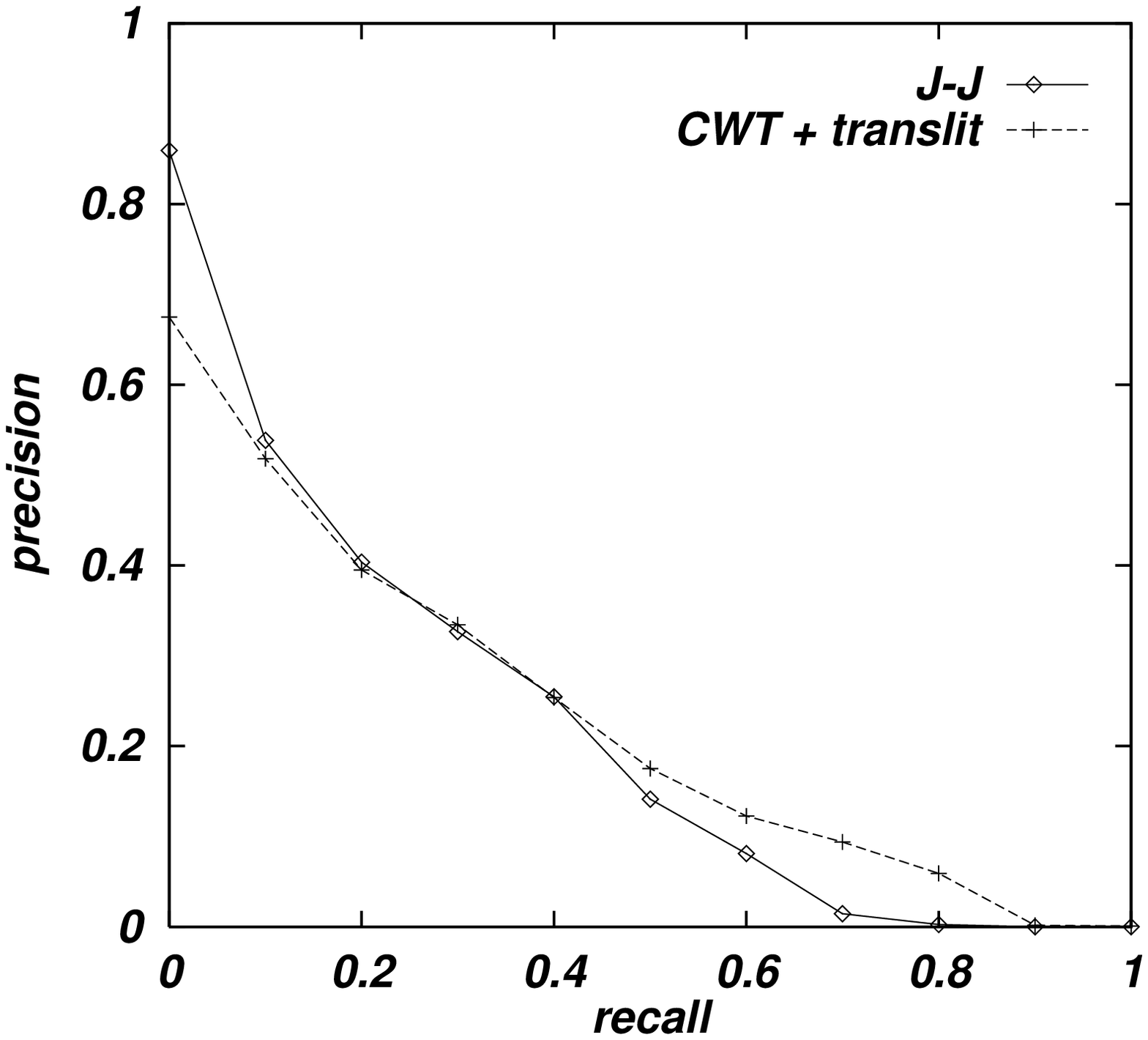,height=2.5in}
  \end{center}
  \caption{Recall-Precision curves for evaluation of overall performance}
  \label{fig:rp_overall}
\end{figure}

We compared our system (``CWT+translit'') with the Japanese-Japanese
IR system, where (unlike the evaluation in
Section~\ref{subsec:eval_translit}) transliteration was applied only
to ``{\it ma-i-ni-n-gu\/}~(mining)'' and ``{\it
ko-ro-ke-i-sho-n\/}~(collocation)''.  Figure~\ref{fig:rp_overall} and
Table~\ref{tab:ap_overall} show the recall-precision curve and
11-point average precision for each system, respectively, from which
one can see that our CLIR system is quite comparable with the
monolingual IR system in performance.  In addition, from
Figure~\ref{fig:rp_cwt} to \ref{fig:rp_overall}, one can see that the
monolingual system generally performs better at lower recall while the
CLIR system performs better at higher recall.

For further investigation, let us discuss similar experimental results
reported by Kando and Aizawa~\shortcite{kando:iral-98}, where a
bilingual dictionary produced from Japanese/English keyword pairs in
the NACSIS documents is used for query translation.  Their evaluation
method is almost the same as performed in our experiments. One
difference is that they use the ``OpenText'' search
engine\footnote{Developed by OpenText Corp.}, and thus the performance
for Japanese-Japanese IR is higher than obtained in our
evaluation. However, the performance of their Japanese-English CLIR
systems, which is roughly 50-60\% of that for {\em their\/}
Japanese-Japanese IR system, is comparable with our CLIR system
performance.  It is expected that using a more sophisticated search
engine, our CLIR system will achieve a higher performance than that
obtained by Kando and Aizawa.

\begin{table}[htbp]
  \def\baselinestretch{1}
  \begin{center}
    \caption{Comparison of average precision for evaluation of overall
    performance}
    \medskip
    \leavevmode
    \small
    \begin{tabular}{|c||c|c|} \hline
      & avg. precision & ratio to J-J \\ \hline\hline
      J-J & 0.204 & --- \\ \hline
      CWT + translit & 0.212 & 1.04 \\ \hline
    \end{tabular}
    \label{tab:ap_overall}
  \end{center}
\end{table}

\section{Conclusion}
\label{sec:conclusion}

In this paper, we proposed a Japanese/English cross-language
information retrieval system, targeting technical documents.  We
combined a query translation module, which performs compound word
translation and transliteration, with an existing monolingual
retrieval method. Our experimental results showed that compound word
translation and transliteration methods individually improve on the
baseline performance, and when used together the improvement is even
greater.  Future work will include the application of automatic word
alignment methods~\cite{fung:acl-95,smadja:cl-96} to enhance the
dictionary.

\section*{Acknowledgments}

The authors would like to thank Noriko Kando (National Center for
Science Information Systems, Japan) for her support with the NACSIS
collection.

\end{document}